\documentclass[10pt,a4paper]{article}

\usepackage[T1]{fontenc}
\usepackage[utf8]{inputenc}
\usepackage{authblk} % For author affiliations
\usepackage[margin=1in]{geometry}
\usepackage{graphicx}
\usepackage{booktabs} % For professional tables
\usepackage{amsmath,amssymb}
\usepackage{url}
\usepackage{natbib}
\usepackage{xcolor}

\usepackage{microtype}
\usepackage{placeins} % For \FloatBarrier
\usepackage{titlesec} % For custom section titles
\usepackage{multirow}
\usepackage{float}

\usepackage{hyperref}
\usepackage{array}
\usepackage{float}
\usepackage{algorithm}
\usepackage{algpseudocode}

\usepackage{hyperref}       % hyperlinks

\usepackage{amsfonts}       % blackboard math symbols
\usepackage{nicefrac}       % compact symbols for 1/2, etc.
\usepackage{microtype}      % microtypography
\usepackage{xcolor}         % colors
\usepackage{amsthm} 
\usepackage{cleveref}
\usepackage{placeins}

\titleformat{\section}
  {\normalfont\Large\bfseries\raggedright}{\thesection.}{1em}{}
\titleformat{\subsection}
  {\normalfont\large\bfseries\raggedright}{\thesubsection.}{1em}{} 
\titleformat{\subsubsection}
  {\normalfont\normalsize\bfseries\raggedright}{\thesubsubsection.}{1em}{} 
\usepackage{hyperref}
\hypersetup{
    colorlinks=true,
    linkcolor=blue,
    filecolor=magenta,      
    urlcolor=blue,
    citecolor=blue,
    pdftitle={Beyond the Parameter Monolith: Reconstructive Memories, Executable Skills, and Residual Assembly for Language Models},
    pdfauthor={Andrey Bochkov},
}

\newcommand{\method}{FEM-ASM}
\newcommand{\R}{\mathbb{R}}

\algrenewcommand\algorithmicrequire{\textbf{Input:}}
\algrenewcommand\algorithmicensure{\textbf{Output:}}

\title{Beyond the Parameter Monolith: Reconstructive Memories, Executable Skills, and Residual Assembly for Language Models}

\author[1]{Andrey Bochkov\thanks{Corresponding author: andrey.bochkov@gmail.com}}

\date{} % To remove the date

\begin{document}

\maketitle

\begin{abstract}
Can a language-model system grow its external state and use
executable procedures without making every component part of one
shared optimization process?
We present \method{}, an implemented architecture that separates
neural coordination, document-state construction, and deterministic
execution.
Independent workers optimize document memories under a frozen
codec and publish immutable, versioned artifacts.
Executable skills return bounded computation results.
Both enter a shared neural field through type-specific adapters,
explicit support mappings, and a common residual-assembly operator.
The resulting computation couples a quadratic consistency substep
to a recurrent learned correction.
We evaluate memory construction, controlled retrieval, positional
readout, execution-result rendering, and evidence attachment with
unchanged core weights.
The experiments show that recoverable storage, successful retrieval,
and successful neural use are distinct properties.
Position-specific observations improve the tested memory and
execution-result interfaces, while fixed-core attachment demonstrates
that newly available evidence can affect predictions through the
existing interface.
The contribution is an architectural and engineering account of
independently produced state participating in shared neural
computation, with explicit component contracts, executable protocols,
and released research artifacts.
\end{abstract}

\section{Introduction}
\label{sec:introduction}

A language model, a document archive, and an executable procedure
differ in what must be learned, what can be verified, and what must
change when new information arrives.
A document update changes stored evidence.
An arithmetic procedure specifies an operation with exact execution
semantics.
A neural core learns how to interpret a request, combine available
information, and produce language.
Treating these roles as separate components creates an architectural
opportunity: their construction and update schedules can be
decoupled while their contributions remain integrated at inference.

Retrieval-augmented and tool-using systems already establish the
usefulness of external resources
\citep{lewis2020rag,borgeaud2022retro,schick2023toolformer}.
The question pursued here concerns the integration boundary:
\emph{how can independently constructed states and execution
results participate directly in a shared neural computation?}

We introduce \method{}, an attention-free modular architecture
with three interacting parts.
A reconstructive-memory pipeline locally optimizes document states
under a frozen shared codec.
A deterministic virtual machine executes bounded programs.
A neural coordination core receives both through typed observations,
explicit target supports, and normalized residual assembly.
The external components retain their own construction procedures;
the core learns their shared use through type-level adapters.

\paragraph{Why independent construction matters.}
The architectural value is the separation of three processes that
otherwise become entangled: optimizing a component, publishing
its state, and consuming that state.
Workers can construct different document artifacts without
synchronizing gradients with one another or with the core.
A registry publishes immutable versions, and consumers use a fixed
snapshot rather than a changing live collection.
An executor produces a result without differentiating through its
instructions.
These independently produced objects become inputs to the same
neural update rule.
The growing object is the external state collection, while the
number of type-shared interface parameter sets remains fixed.
This organization makes component-level construction, replacement,
replay, and inspection explicit architectural operations.

The design draws on domain decomposition and finite-element
assembly \citep{toselli2005domain}.
Its transferable principle is that independently computed local
objects interact through specified supports and a common assembly
rule.
In \method{}, the objects are document and execution observations,
the supports are neural sequence nodes, and assembly reconciles
their proposals in the model's state space.

The central question is not whether tools or retrieval can help
a language model.
It is whether a common internal interface can coordinate
heterogeneous external objects while preserving their independent
lifecycles, and which interface choices make their information usable.

We investigate this through connected experiments.
A reconstructed cell may retain source information but make a poor
retrieval key.
A retriever may locate the correct block while answer generation
still fails.
An executor may compute the correct integer while its neural
rendering is wrong.
These observations motivate separating \emph{storage},
\emph{addressability}, \emph{readout}, and \emph{rendering}.
The resulting interface uses component-specific observations
with explicit positional supports rather than relying solely on
repeated global vectors.

Our contributions are:
\begin{enumerate}
    \item a typed residual interface connecting independently
    constructed memories and executable results to a shared
    recurrent neural field;
    \item an asynchronous memory-construction and versioned
    publication pipeline with snapshot-based consumption;
    \item controlled interface experiments and executable
    algorithms identifying where external information becomes
    usable and how failures can be localized.
\end{enumerate}

\section{Related Work}
\label{sec:related}

\paragraph{External resources.}
RAG, kNN language models, RETRO, and Memorizing Transformers
augment neural computation with external storage
\citep{lewis2020rag,khandelwal2020knnlm,
borgeaud2022retro,wu2022memorizing}.
Toolformer studies learned access to executable APIs
\citep{schick2023toolformer}.
\method{} investigates the combination of locally optimized
document states, versioned publication, explicit support mappings,
and a common internal return interface for memory and execution.
Its experimental focus is the path from an independently
produced object to its effect on neural output.

\paragraph{Shared computation and assembly.}
Universal Transformers and deep equilibrium models provide
precedents for recurrent and shared computation
\citep{dehghani2019universal,bai2019deq}.
Permutation-invariant aggregation is established by Deep Sets
\citep{zaheer2017deepsets}.
Domain decomposition supplies the architectural inspiration for
local computation and interface reconciliation
\citep{toselli2005domain}.
The contribution here is their concrete organization into a
typed external-state system with separate production and
consumption protocols.

\paragraph{Sequence and input representations.}
Transformers, Mamba, and Hyena provide different sequence
operators
\citep{vaswani2017attention,gu2023mamba,poli2023hyena};
Compressive Transformers study multiresolution history
\citep{rae2020compressive}.
Our earlier Multi-Mesh prototype supplies the local,
multiresolution background for the modular core.
Fixed Binary16 inputs and optional visual-form observations
connect to work on frozen Unicode representations and
pixel-based language modeling
\citep{bochkov2025emergent,rust2023pixel}.
The present contribution concerns component coordination above
the input interface.

\section{Architecture: Independent Producers, Shared Computation}
\label{sec:architecture}

\begin{figure}[t]
\centering
\includegraphics[width=0.78\linewidth]{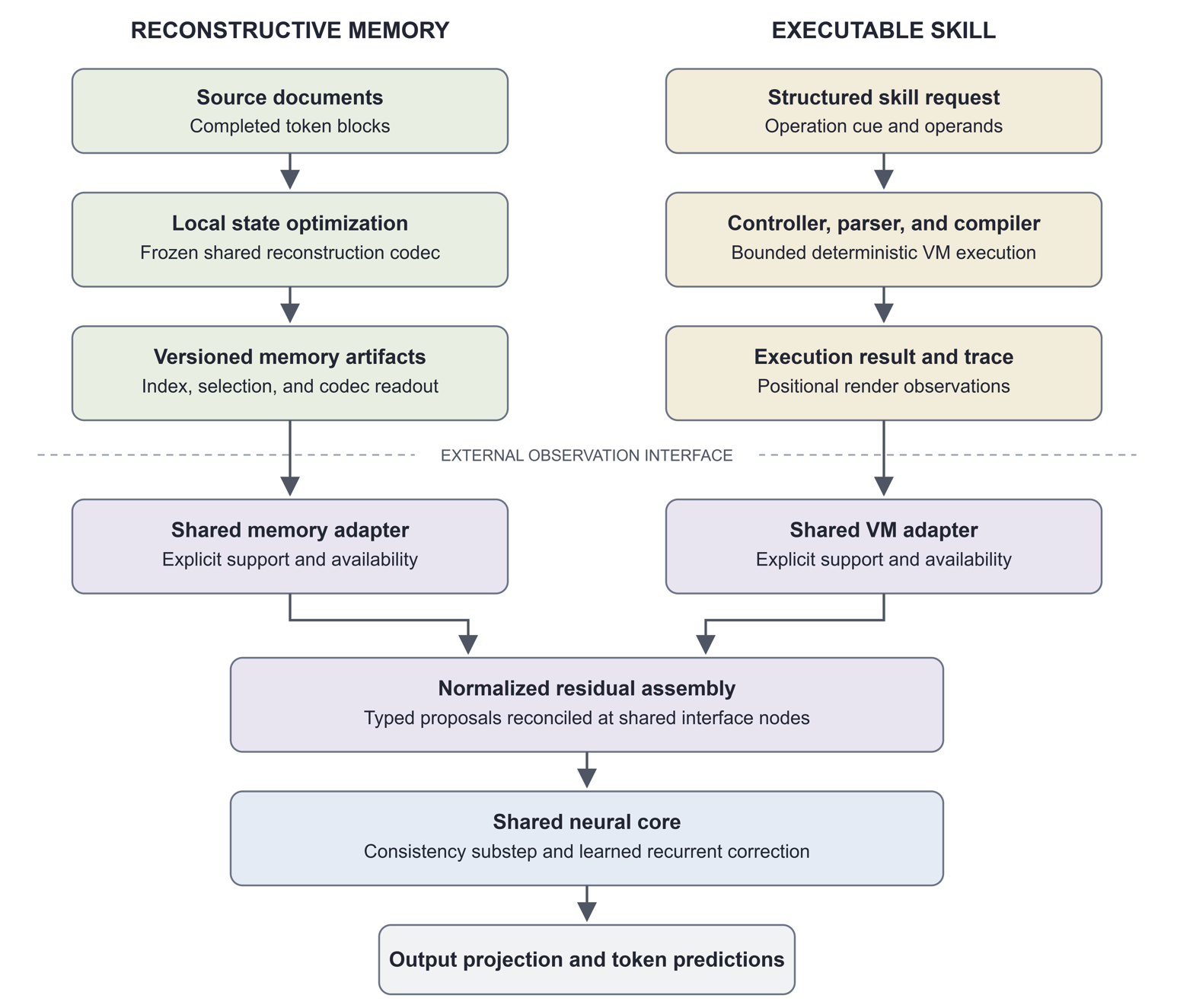}
\caption{
Independent construction and shared neural consumption.
Memory workers publish versioned document states under a frozen codec.
The executable path returns VM results and rendered observations.
Type-shared adapters map both to supported proposals in the same
neural field.
Memory selection and positional alignment follow the specified
evaluation protocol.
}
\label{fig:architecture}
\end{figure}

Figure~\ref{fig:architecture} summarizes the system.
Document workers and executable skills produce observations through
different paths.
They meet at a common adapter--support--assembly interface.

\subsection{Typed observations}

Let $Z_{b,i}^k\in\R^d$ denote the state of sample $b$ at node $i$
and iteration $k$.
An observation specifies
$$
e=(\tau_e,u_e,b_e,i_e,a_e,w_e,p_e),
$$
where $\tau_e$ is its type, $u_e$ its native state,
$b_e$ its sample, $i_e$ its target node, $a_e$ its availability
boundary, $w_e\geq0$ its weight, and $p_e$ its provenance.

A shared type adapter produces
$$
z_e=T_{\tau_e}(u_e).
$$
No adapter is indexed by document identity.
Provenance is retained for diagnostics and artifact management.
A component may emit multiple observations, such as reconstructed
slots or successive rendered result tokens.

The implementation validates dimensions, finite values, weights,
indices, and the attachment condition $i_e\geq a_e$.
Source construction and support selection remain explicit parts
of the interface contract.

\subsection{Residual assembly and recurrent processing}

For observations attached to node $(b,i)$, define
$$
R_{b,i}^k=
\frac{\sum_{e\rightarrow(b,i)}w_e(z_e-Z_{b,i}^k)}
     {\epsilon+\sum_{e\rightarrow(b,i)}w_e}.
$$
Segment sums implement this operator with FP32 accumulation.
Unconstrained nodes receive zero external residual.
The mathematical sum is permutation invariant.

The update separates external reconciliation from learned dynamics:
$$
Z_{\mathrm c,b}^k=Z_b^k+\alpha_b^k R_b^k,
$$
$$
Z_b^{k+1}=Z_{\mathrm c,b}^k+
\beta P_\theta\!\left(
Z_{\mathrm c,b}^k,\,
\lambda N_{\mathrm l}(S_\theta(Z_b^k))
+N_{\mathrm e}(R_b^k)
\right).
$$
Here $S_\theta$ is a shared causal local operator,
$N_{\mathrm l}$ and $N_{\mathrm e}$ normalize messages,
and $P_\theta$ is a shared correction operator with bounded
$\tanh$ output.
The supplied implementation uses $\beta=s/4$, where $s$ is
the configured update scale.

For fixed proposals and
$$
\mathcal{E}(Z)=\frac12\sum_e
w_e\lVert z_e-Z_{b_e,i_e}\rVert^2,
$$
any consistency step with $0\leq\alpha_b^k\leq1$ cannot
increase the corresponding sample energy in exact arithmetic.
The implementation chooses a scale from a fixed backtracking
sequence using a per-sample energy check.
The learned correction is optimized for the language objective
and is not subject to that acceptance test.

Appendix~\ref{app:algorithms} gives the implemented update;
Appendix~\ref{app:energy} derives its consistency property and
distinguishes the analytic local update from the numerical
sample-level safeguard.

\subsection{Document construction and publication}

A document is divided into completed blocks of 64 tokens.
The frozen codec produces initial cells $C^0$.
A worker optimizes document-specific residuals $M$ and forms
$$
C=C^0+M.
$$
The decoder predicts source positions from cells, positional
coordinates, and optional parent context, without receiving
the correct preceding tokens of the reconstructed block.

Workers share codec weights but not document-state gradients.
Each produces its own adaptation record and publication decision.
Accepted artifacts contain final cells, derived coarse states,
source checksums, and metadata.
Publication and activation are separate registry operations.

The registry maintains an append-only event journal.
Consumers use immutable snapshots containing exact element
versions and checksums.
New publication therefore does not change the evidence set
of an ongoing snapshot-based run.
Algorithms~\ref{alg:worker}--\ref{alg:snapshot} specify these
boundaries.

\subsection{Distributed construction and compatibility}

The system distributes artifact production rather than requiring
a distributed neural forward pass.
Each worker receives document tasks and a frozen codec version,
performs local optimization, and publishes completed artifacts.
The coordination point is the registry's publication protocol,
not synchronized optimization of all document states.
A consumer may load a snapshot while producers continue working
on elements for a later snapshot.

Compatibility has both structural and learned parts.
Tensor dimensions, codec identity, tokenization, slot ordering,
and availability rules form the structural contract.
The adapter and core must also have learned to use observations
from the relevant representation family.
Adding another document under an established codec preserves
this interface family.
Replacing the codec itself can change the latent coordinates
and requires explicit compatibility testing or adaptation.

This distinction is central to the architecture.
An immutable file is not automatically an interchangeable neural
module.
The registry identifies what was published; the interface contract
specifies how it can be consumed; evaluation establishes whether
the consumer uses it successfully.

\subsection{Retrieval, readout, and executable return}

Memory addressability is supplied by prefix, sliding-window,
token-bag, and $n$-gram sketches, followed by a shared reranker.
Selected content is exposed either as a cell vector or through
position-specific codec states.
Retrieval geometry and reconstruction geometry can therefore
be changed independently.

The skill path selects an operation, parses operands, compiles
a bounded program, executes it, and converts the result into
observations.
A second neural evaluation consumes the returned observations.
This closes the information path from neural routing through
external execution back to neural output; VM execution remains
outside autograd.

The successful arithmetic protocol learns operation selection
among five operations.
Its positional return buffer includes rendered token identities
and uses the externally known output length.
Memory diagnostics specify separately whether source positions
and answer alignment are supplied by the task record.
The integrated dataflow is given in Algorithm~\ref{alg:cycle}.

\begin{figure}[t]
\centering
\includegraphics[width=\linewidth]{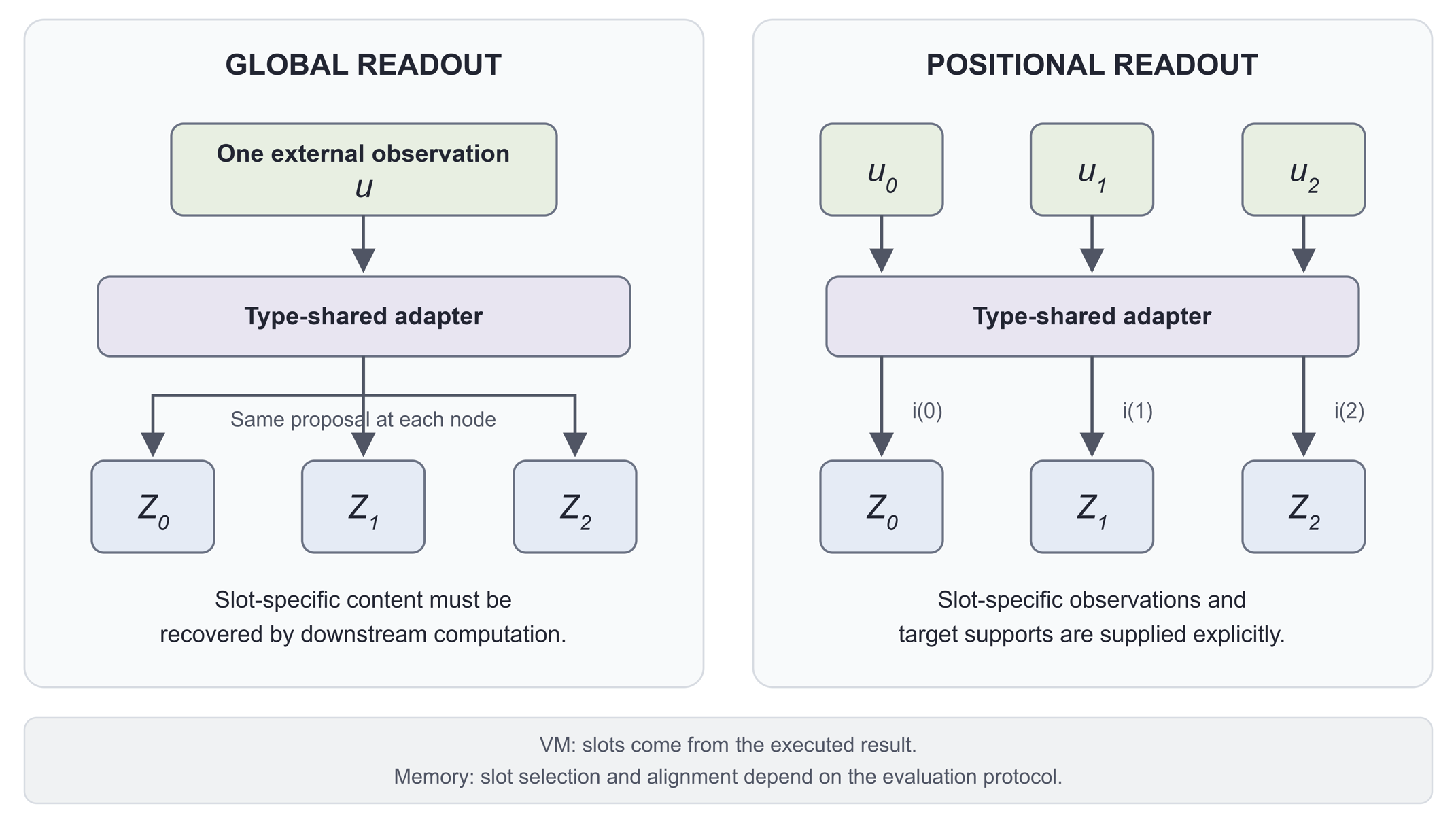}
\caption{
Global and positional external readout.
A repeated global observation leaves slot-specific content recovery
to downstream computation.
A positional interface exposes distinct observations with explicit
supports.
VM slots are constructed from the executed result; memory selection
and alignment follow the declared task protocol.
The compared interfaces differ in supplied features as well as
positional organization.
}
\label{fig:readout}
\end{figure}

\section{Experimental Protocol}
\label{sec:protocol}

\paragraph{Data and conditioning.}
Experiments use FineWeb-Edu \citep{penedo2024fineweb}
and a SmolLM2-family tokenizer \citep{allal2025smollm2}.
Memory workers use designated odd training shards; the base
modular language stream uses even shards.
QA examples are split by element identity before generation.
Memory-conditioned supervision also exposes source-document
information to the core.

The QA tasks include prefix completion, masked-span restoration,
value restoration, and quote continuation.
Several diagnostics construct cues from source spans and stored
offsets.
These are coordinate-assisted addressability and readout
experiments.
The attachment package supplies selected evidence directly.

\paragraph{Controls and metrics.}
Memory controls use absent, wrong, and permuted observations.
Execution-result interventions substitute the returned value.
Attachment evaluations keep core weights unchanged.
We report token NLL, token accuracy, and answer metrics separately.
VM numeric exact match extracts the first signed integer and
uses the buffer's declared generation length.

\paragraph{Experimental models.}
The standalone Multi-Mesh prototype and the later modular
core are distinct architectures.
The former characterizes attention-free multiresolution language
modeling.
The latter tests external-element adaptation and assembly.
Configurations and implementation boundaries are documented in
Appendix~\ref{app:training}.

\section{Results}
\label{sec:results}

\subsection{Standalone Multi-Mesh and a same-corpus Transformer}

The Multi-Mesh prototype combines a fine token field, structured
binary tiles, completed-span coarse fields, and a scratch field.
We evaluate it without external memory or VM.

As a conventional reference, we include AB-EXT Learned,
a 24-layer decoder-only Transformer with 1.812B trainable
parameters, trained from scratch on the same FineWeb-Edu
data source with a 100B-token budget and the same tokenizer
family.
The comparison measures attained checkpoint quality;
training budgets are not matched.

\begin{table}[t]
\centering
\small
\caption{
Standalone language-model evaluation.
The Transformer is a same-corpus reference with a 100B-token
training budget, not a matched-budget control.
External memory and VM are disabled.
Accuracy entries are percentages.
}
\label{tab:standard}
\begin{tabular}{lrr}
\toprule
Metric & Multi-Mesh & Transformer \\
\midrule
HellaSwag accuracy & 29.00 & 44.21 \\
HellaSwag normalized accuracy & 30.77 & 57.79 \\
ARC-Easy accuracy & 53.37 & 71.63 \\
ARC-Challenge accuracy & 21.59 & 35.92 \\
PIQA accuracy & 62.95 & 72.69 \\
WinoGrande accuracy & 49.33 & 58.56 \\
MMLU 0-shot accuracy & 23.95 & 25.32 \\
LAMBADA accuracy & 4.87 & 47.72 \\
WikiText word perplexity $\downarrow$ & 65.02 & 16.50 \\
WikiText bits per byte $\downarrow$ & 1.13 & 0.76 \\
\bottomrule
\end{tabular}
\end{table}

The tasks follow HellaSwag, ARC, PIQA, WinoGrande, MMLU,
LAMBADA, and WikiText
\citep{zellers2019hellaswag,clark2018arc,bisk2020piqa,
sakaguchi2020winogrande,hendrycks2021mmlu,
paperno2016lambada,merity2017pointer}.

The Transformer obtains stronger standalone scores on most
tasks.
Multi-Mesh learns causal language modeling through its local
and multiresolution operators.
The following experiments concern the distinct modular core
and the use of external observations.

\subsection{Independent memory production}

The production run published 52,809 active element versions
near a budget of 1.7 billion stored floating-point values.
An element can contain multiple block cells.

A selected-cache audit covered 2,910 blocks from 98 elements:
139,468 of 186,031 source tokens were reconstructed exactly,
giving 74.97\% token accuracy.
One complete block was reconstructed exactly.
The results establish recoverable source information while
distinguishing token recovery from complete-span fidelity.
Artifact counts, storage, and the cache schema appear in
Appendix~\ref{app:memory}.

\subsection{Addressability does not determine answer quality}

On the full-S10 QA dataset of 533,978 tasks from 5,296 elements,
the multi-chart generator reached approximately 99.6\% test
candidate recall and the reranker approximately 97.7\% recall
at one under the coordinate-assisted lexical protocol.

Answer-token NLL was 6.6930 with retrieved evidence,
6.7726 with wrong evidence, and 7.2544 without evidence.
Correct content therefore supplies an additional benefit beyond
generic external conditioning.
The same run had zero measured greedy answer exact match.
Retrieval success and likelihood improvement were insufficient
for free answer generation in that configuration.

This outcome separates two engineering tasks.
The index determines which block enters the interface.
The readout determines which information from that block is
available at each receiving node.
Improving the first does not automatically solve the second.

\subsection{Positional observations improve the tested readouts}

Table~\ref{tab:interfaces} brings together the memory and
execution-result interface comparisons.
Memory measurements come from a developmental readout diagnostic,
not the full-S10 retrieval run.
VM measurements use 128 generation examples per range.

\begin{table}[t]
\centering
\small
\caption{
External readout comparisons.
Memory accuracy is teacher-forced token accuracy.
VM scores are numeric exact match under buffer-length-controlled
generation; the wider-range interval includes the training range.
The interfaces differ in supplied features as well as support.
}
\label{tab:interfaces}
\begin{tabular}{lrr}
\toprule
Memory interface & NLL & Token accuracy (\%) \\
\midrule
Repeated dense cell & 6.6331 & 12.15 \\
Source positional & 5.5522 & 30.32 \\
Codec positional & 4.4729 & 48.60 \\
Oracle answer buffer & 4.2934 & 55.41 \\
\midrule
VM return interface & In-range (\%) & Wider-range (\%) \\
\midrule
Repeated global vector & 8.59 & 6.25 \\
Character positional & 88.28 & 35.94 \\
Tokenizer-aware positional & 100.00 & 85.94 \\
Token + character positional & 100.00 & 88.28 \\
\bottomrule
\end{tabular}
\end{table}

For memory, codec slots improve NLL by more than two nats
relative to the dense-cell interface.
Wrong-memory and slot-permutation interventions increase the
positional model's loss, indicating dependence on both content
and slot organization.
The oracle answer-buffer result identifies remaining error
in the neural readout path even when target information is supplied.

For arithmetic, routing and deterministic execution reach
100\% on the reported test sets, while rendering depends
strongly on the return interface.
The execution result is already available; the experiment
isolates its neural adoption and serialization.
Substituted returns change answer likelihood and can redirect
the generated answer.

The shared lesson is a division of computational responsibility.
A global observation asks the receiving network to recover
content, position, and serialization structure together.
Component-specific readout can expose that structure before
assembly.
The codec supplies slot-wise reconstruction states; the VM
renderer supplies an ordered representation of an exact result.
The shared core then learns to use these observations in context.
Additional controls and scoring details are given in
Appendices~\ref{app:readout} and~\ref{app:vm}.

\subsection{Attachment with unchanged core weights}

The packaged attachment demonstration evaluates 417 examples
and 2,396 target tokens.
Required evidence is absent in the first condition and
available in the second.

\begin{table}[t]
\centering
\small
\caption{
Selected-evidence attachment with fixed core weights.
Source labels identify the originating memory conditions;
the package loads selected evidence, not the entire stores.
}
\label{tab:attachment}
\begin{tabular}{lrrr}
\toprule
Source condition & Coverage & NLL & Token accuracy (\%) \\
\midrule
\texttt{G1p7} & 0.00 & 6.5716 & 12.10 \\
\texttt{G6p5} & 1.00 & 2.8073 & 58.72 \\
\bottomrule
\end{tabular}
\end{table}

Evidence attachment reduces NLL by 3.7643 nats per target token.
The core uses the newly available observations through its
existing interface without a parameter update.
This measures evidence availability, with full-store retrieval
and serving costs treated separately.

\section{Architectural Implications}
\label{sec:discussion}

The system separates the representation an external component
uses internally from the observations a neural consumer needs.
Document reconstruction does not require its latent space to
double as a retrieval metric.
Exact execution does not require a neural decoder to infer a
decimal representation from one global vector.
Component-specific readout performs that translation, while
type-shared adapters and assembly provide a common destination.

The distributed aspect lies primarily in construction and
lifecycle management.
Memory optimization is local to each document and can proceed
on independent workers.
Publication serializes artifact visibility, not gradient updates.
Snapshots fix the evidence set consumed by a run.
Neural coordination remains shared, but growth in the number
of document artifacts does not introduce document-specific
adapter parameters.
Retrieval, storage, and readout costs remain functions of the
external collection and selected observations.

This is the organizing contribution of \method{}:
independently produced state is integrated into shared neural
computation through an explicit contract rather than through
a common training history.
The experiments identify practical requirements for that
contract and make failures localizable.

The resulting architecture supports component-wise research.
A new search index can be evaluated against unchanged cells.
A revised readout can be tested with oracle selection before
introducing retrieval errors.
An execution interface can be tested by substituting returns
without changing arithmetic semantics.
A new memory snapshot can be attached while core weights
remain fixed.
Each intervention has a defined boundary, making system
development more interpretable than changing all components
at once.

\section{Limitations}
\label{sec:limitations}

The current memory is lossy and occupies more storage than raw
token IDs: a 768-value BF16 cell uses 1,536 bytes for a
64-token block whose uint16 IDs require 128 bytes.
Parent states, indices, and readout caches add further cost.
Matched raw-text and simpler feature-injection baselines are
needed to isolate the benefit of this particular assembly
architecture.

Several QA protocols use source-derived coordinates or selected
evidence.
Element-ID splitting does not establish content deduplication
or absence of initialization exposure.
Arithmetic uses deterministic parsing and compilation, intended
output-token observations, and external length control.
Some rendering implementations differ between training and
greedy decoding in token feedback and historical attachments.

Most interface measurements are finite-sample point estimates.
The package does not measure complete-store serving or
fixed-answerability scaling.
The implemented sample-level backtracking safeguard can couple
position updates through its acceptance decision; strict
prefix independence requires the conditions described in
Appendix~\ref{app:energy}.
The consistency guarantee concerns the external substep,
not the learned correction or answer correctness.

\section{Conclusion}

\method{} implements a separation between neural coordination,
reconstructive document state, and deterministic execution.
Independent producers create artifacts and results; a common
typed residual interface brings their observations into a
shared neural field.

The experiments distinguish storage, addressability, readout,
and rendering as separate requirements.
They demonstrate usable positional return interfaces and
evidence-dependent improvement with unchanged core weights.
The accompanying algorithms, diagnostics, and artifacts document
a concrete foundation for language-model systems whose external
state and executable components can be constructed and managed
independently of the core.

\section*{Reproducibility Statement}

The release provides model artifacts, a selected-evidence
attachment demonstration, and documented research code.
The appendices specify the main algorithms, source-code
boundaries, configurations, and evaluation conditions.
Packaged paths and historical research components have different
data and artifact dependencies; the release is not a complete
end-to-end reproduction of every historical run.

\section*{Ethics Statement}

The work uses web-derived data and research models subject to
source licenses, privacy obligations, and ordinary language-model
risks.
Checksums establish integrity rather than factual reliability.
The VM instruction set has no host file, network, or shell
operations; this does not constitute a formal security audit.
Online collection would require explicit consent, retention,
deletion, and poisoning policies.

\bibliographystyle{plainnat}
\bibliography{main}

\clearpage
\appendix

\section{Executable Architecture: Algorithms and Contracts}
\label{app:algorithms}

This appendix begins with the system's operational structure.
Algorithms~\ref{alg:assembly}, \ref{alg:solver},
\ref{alg:worker}, \ref{alg:publish}, \ref{alg:snapshot},
and~\ref{alg:vm} summarize the corresponding supplied
implementations.
Algorithms~\ref{alg:cycle} and~\ref{alg:diagnostics} specify
integration and measurement protocols at a higher level.

The implementation separates three boundaries:
\begin{enumerate}
    \item \textbf{Production:} source information becomes a
    memory artifact or an execution result.
    \item \textbf{Publication and selection:} versions and
    observations become available to a consumer.
    \item \textbf{Neural consumption:} validated observations
    become proposals and participate in recurrent state updates.
\end{enumerate}

\subsection{Typed adaptation and matrix-free assembly}

For a batch of $B$ sequences of length $T$, a target pair
$(b_e,i_e)$ is flattened to $n_e=b_eT+i_e$.
All sums are accumulated in FP32.
Adapted proposals are computed once per backbone call and
reused across its iterations.

\begin{algorithm}[H]
\caption{Typed adaptation and external residual assembly}
\label{alg:assembly}
\begin{algorithmic}[1]
\Require State $Z\in\R^{B\times T\times d}$; typed element batches;
type adapters $T_\tau$; weight scale $\gamma$; $\epsilon>0$
\Ensure Adapted proposals $\mathcal A$, residual $R$, node weights $W$
\State $\mathcal A\gets\varnothing$
\For{each typed batch of external observations}
    \State Validate source dimensions, finite states and weights,
    index bounds, and $i_e\geq a_e$
    \State $z_e\gets T_{\tau_e}(u_e)$ for every observation
    \State $n_e\gets b_eT+i_e$; $\widehat w_e\gets\gamma w_e$
    \State Add $(z_e,n_e,\widehat w_e,p_e)$ to $\mathcal A$
\EndFor
\State Flatten $Z$ to $\widehat Z\in\R^{BT\times d}$
\State $S\gets0\in\R^{BT\times d}$; $W\gets0\in\R^{BT}$
\For{each adapted batch in $\mathcal A$}
    \State $S[n_e]\mathrel{+}=
    \widehat w_e(z_e-\widehat Z[n_e])$ using segment sum
    \State $W[n_e]\mathrel{+}=\widehat w_e$ using segment sum
\EndFor
\State $R[n]\gets S[n]/(\epsilon+W[n])$ where $W[n]>0$
\State $R[n]\gets0$ where $W[n]=0$
\State Reshape and cast $R,W$ to the state layout and dtype
\State \Return $\mathcal A,R,W$
\end{algorithmic}
\end{algorithm}

\paragraph{Implemented adapter.}
The supplied adapter normalizes the source and combines gated
and skip projections:
$$
x=N_{\mathrm{src}}(u),
$$
$$
T_\tau(u)=N_{\mathrm{out}}
\left(
\operatorname{SiLU}(W_{\mathrm{in}}x)
\odot\sigma(W_{\mathrm{gate}}x)
+W_{\mathrm{skip}}x
\right).
$$
All projections are shared across instances of the type.
The adapter does not consume provenance IDs.

\subsection{Recurrent reconciliation}

Let $s>0$ be the configured update scale, $J$ the number of
backtracking candidates, and $\delta$ the numerical tolerance.
The implementation evaluates
$$
\alpha_j=\min(1,s/2^j),
\qquad j=0,\ldots,J-1.
$$
Each sample accepts the first candidate that passes its
energy test.

\begin{algorithm}[H]
\caption{Implemented recurrent residual solver}
\label{alg:solver}
\begin{algorithmic}[1]
\Require Token IDs $x$; external observations; iterations $K$;
update scale $s$; backtracking count $J$; tolerance $\delta$
\Ensure Final hidden states and optional controller outputs
\State $Z\gets N_{\mathrm{in}}(\operatorname{Binary16Lift}(x))$
\State Validate and adapt observations once to obtain $\mathcal A$
\State $E_b\gets\operatorname{Energy}_b(Z,\mathcal A)$
\For{$k=0,\ldots,K-1$}
    \State $L\gets S_\theta(Z)$
    \State $(R,W)\gets\operatorname{Assemble}(Z,\mathcal A)$
    \State $Z_{\mathrm c}\gets Z$; mark all samples unaccepted
    \For{$j=0,\ldots,J-1$}
        \State $\alpha_j\gets\min(1,s/2^j)$
        \State $Y\gets Z+\alpha_jR$
        \State $E_b^{\mathrm{cand}}\gets
        \operatorname{Energy}_b(Y,\mathcal A)$
        \For{each unaccepted sample $b$ with positive external weight}
            \If{$E_b^{\mathrm{cand}}\leq E_b+\delta$}
                \State $Z_{\mathrm c,b}\gets Y_b$;
                record $\alpha_b=\alpha_j$; mark accepted
            \EndIf
        \EndFor
    \EndFor
    \State $Q\gets\lambda N_{\mathrm l}(L)+N_{\mathrm e}(R)$
    \State $\Delta_{\mathrm n}\gets(s/4)P_\theta(Z_{\mathrm c},Q)$
    \Comment{$P_\theta$ ends with $\tanh$}
    \State $Z_{\mathrm{new}}\gets Z_{\mathrm c}+\Delta_{\mathrm n}$
    \State Record energies before, after consistency, and after
    the full update when tracing is enabled
    \State $Z\gets Z_{\mathrm{new}}$;
    $E_b\gets\operatorname{Energy}_b(Z,\mathcal A)$
\EndFor
\State $H\gets N_{\mathrm{final}}(Z)$
\State \Return $H$ and requested VM-controller heads
\end{algorithmic}
\end{algorithm}

The local and external residuals used by the correction are
computed from the pre-consistency state.
The correction receives the post-consistency state.
The next iteration measures energy from the actual
post-correction state.

The output projection is trainable and untied.
The training wrapper consumes already-shifted labels:
inputs are the first $T$ tokens and targets the following
$T$ tokens.
Weighted joint losses use supplied weights as a sum;
the trainer determines their task-level normalization.

\subsection{Independent document workers}

\begin{algorithm}[H]
\caption{Asynchronous document-state construction}
\label{alg:worker}
\begin{algorithmic}[1]
\Require Assigned document tasks; frozen codec $\phi$;
adaptation settings; quality thresholds; registry; stop path
\Ensure Accepted element versions published independently
\State Load codec configuration and weights; freeze all codec parameters
\For{each assigned document task}
    \If{the stop path exists}
        \State Stop before starting the next document
    \EndIf
    \If{the element version already has a READY event}
        \State Skip this task
    \Else
        \State Load source tokens and validate their count
        \State Prepare fixed observations and initial codec cells $C^0$
        \State Adapt only document residuals $M$ under frozen $\phi$
        \State Form $C=C^0+M$ and obtain reconstruction metrics
        \State Save a local adaptation record:
        $C^0,M,C$, metrics, and adaptation curve
        \If{recall and reconstruction-loss thresholds are satisfied}
            \State Cast final cells to BF16 for publication
            \State Build coarse states and source-token checksum
            \State Write artifact with codec identity and metadata
            \State Publish through Algorithm~\ref{alg:publish}
            \State Optionally remove the local adaptation record
        \EndIf
    \EndIf
    \State Record task failures according to the worker's error policy
\EndFor
\end{algorithmic}
\end{algorithm}

The worker invokes the shared adaptation routine rather than
updating the codec.
The published file contains final cells, configured coarse
states, and a source checksum.
Residual and initial-cell tensors belong to the local adaptation
record and are not necessarily retained in the registry artifact.

The worker applies its quality gate to adaptation metrics before
publishing the BF16 representation.
Post-serialization reconstruction is a separate measurement,
exemplified by the readout-cache audit.

\subsection{Publication and snapshot isolation}

\begin{algorithm}[H]
\caption{Immutable element publication}
\label{alg:publish}
\begin{algorithmic}[1]
\Require Initialized registry; completed worker file;
element ID, version, codec ID, provenance; optional activation
\Ensure READY event referencing a published artifact
\State Validate IDs, metadata, source existence, and file suffix
\State Copy source to a unique temporary file beside the destination
\State Flush and synchronize the temporary file
\State Compute SHA-256 and physical byte count
\State Verify an expected checksum if supplied
\State Acquire the registry's exclusive lock
\State Replay the journal to materialize current state
\State Reject duplicate published versions or an occupied destination
\State Atomically rename the temporary file to the final path
\State Synchronize the destination directory
\State Append and synchronize READY with path, checksum, codec, and provenance
\If{activation was requested}
    \State Append and synchronize ACTIVATE for this version
\EndIf
\State Release the registry lock
\State \Return the publication descriptor
\end{algorithmic}
\end{algorithm}

\begin{algorithm}[H]
\caption{Snapshot creation and fixed-snapshot consumption}
\label{alg:snapshot}
\begin{algorithmic}[1]
\Require Registry; optional codec filter; file-verification policy
\Ensure Snapshot identifying an exact active set
\State Acquire the registry's exclusive lock
\State Replay READY, ACTIVATE, DEACTIVATE, and RETIRE events
\State Select active, non-retired versions in sorted element-ID order
\State Apply the codec filter when specified
\If{file verification is enabled}
    \State Verify selected files' presence, sizes, and checksums
\EndIf
\State Construct manifest with versions, paths, hashes, provenance,
event count, timestamp, and metadata
\State Compute snapshot ID from canonical serialization of the manifest
\State Write snapshot atomically and append a SNAPSHOT event
\State Release the lock
\State Consumers retain this manifest for the run or declared run segment
\State \Return snapshot path and manifest
\end{algorithmic}
\end{algorithm}

The registry serializes publication and lifecycle events;
document optimization remains outside the lock.
Snapshot creation can verify files while holding the lock.
A snapshot ID identifies its complete manifest, including
timestamp metadata, rather than only the set of active elements.

\subsection{Neural--external--neural information cycle}

\begin{algorithm}[H]
\caption{External selection or execution followed by neural consumption}
\label{alg:cycle}
\begin{algorithmic}[1]
\Require Request $q$; fixed core parameters $\theta$;
memory snapshot; retrieval/readout policy; VM configuration
\Ensure Predictions conditioned on selected external observations
\State Initialize observation collection $\mathcal O$
\If{the protocol requests memory evidence}
    \State Construct retrieval cue under the declared protocol
    \State Search prefix, window, bag, and $n$-gram charts
    \State Merge candidates and apply the shared reranker
    \State Load selected cells and required parent states
    \State Decode slot states or load their readout cache
    \State Construct supports and availability boundaries
    \State Add memory observations to $\mathcal O$
\EndIf
\If{the protocol requests arithmetic execution}
    \State Evaluate the controller on the routing context
    \State Select an arithmetic operation and parse operands
    \State Compile the fixed instruction skeleton
    \State Execute with Algorithm~\ref{alg:vm}
    \If{the result status is HALTED}
        \State Tokenize the rendered result
        \State Construct positional observations and supports
        \State Add VM observations to $\mathcal O$
    \Else
        \State Apply the protocol's execution-failure policy
    \EndIf
\EndIf
\State Evaluate the core with $\mathcal O$ using Algorithm~\ref{alg:solver}
\State Score targets or generate under the declared output protocol
\State \Return predictions and associated provenance
\end{algorithmic}
\end{algorithm}

This is the integration dataflow, not a claim that every
experiment activates both branches.
Coordinate-assisted memory tasks and question-only retrieval
differ at cue and support construction.
The supplied base trainer implements R0, oracle-memory R1,
and oracle-program S1.1.
Later retrieval and execution-result experiments use additional
trainers.

The backbone initializes its field from token coordinates
on each call.
A second reconciliation evaluation therefore need not continue
the previous latent tensor in place.
The closed loop is an information path from neural routing
to external execution and back to neural output.

\subsection{Bounded VM execution}

\begin{algorithm}[H]
\caption{Deterministic bounded integer VM}
\label{alg:vm}
\begin{algorithmic}[1]
\Require Schema-conforming program $P$ and validated VM configuration
\Ensure Typed result, final state, bounded trace, and result checksum
\State Serialize $P$ canonically and compute its program checksum
\State Check program length and initial register/memory counts
\State Initialize registers and memory with signed-width wrapping
\State $pc\gets0$; $steps\gets0$; comparison flag $\gets\mathrm{EQ}$
\State Output and trace $\gets$ empty
\While{no terminal status has been set}
    \If{$steps$ reaches the configured limit}
        \State Set STEP\_LIMIT and exit
    \EndIf
    \If{$pc$ is outside the program}
        \State Set PC\_OOB and exit
    \EndIf
    \State Save pre-instruction register, flag, and PC state
    \State Decode opcode and validate its argument count
    \State Check applicable register, memory, jump, shift, and output constraints
    \State Execute the instruction using specified integer semantics
    \State On a VM fault, set its typed status and message
    \State Increment $steps$ and append the instruction trace
\EndWhile
\State Serialize status, machine state, output, trace, and program checksum
\State Compute result SHA-256
\State \Return the result object
\end{algorithmic}
\end{algorithm}

Validation is performed at the relevant instruction branch.
Input parsing and configuration validation are separate API
boundaries from typed runtime faults.

\subsection{Component diagnostics}

\begin{algorithm}[H]
\caption{Paired diagnostics for external-information use}
\label{alg:diagnostics}
\begin{algorithmic}[1]
\Require Fixed checkpoint; evaluation records;
declared conditioning and scoring protocol
\Ensure Separate storage, retrieval, readout, and rendering measurements
\State Measure codec token and exact-span reconstruction
\State Measure candidate recall before any oracle insertion
\State Measure reranker recall on the declared candidate set
\For{each memory evaluation record}
    \State Evaluate correct and absent observations on identical targets
    \State Evaluate wrong-content and slot-permuted observations
    \State Record likelihood and generation metrics separately
\EndFor
\For{each executable-result record}
    \State Verify routing and VM status independently of rendering
    \State Evaluate the returned-result interface
    \State Substitute a result differing from the original
    \State Evaluate with the same scoring rule
\EndFor
\State Compare attachment conditions with unchanged core weights
\State Retain example, block, and element identities for paired analysis
\State \Return stage-specific metrics and intervention differences
\end{algorithmic}
\end{algorithm}

\section{Consistency, Numerical Acceptance, and Causality}
\label{app:energy}

\subsection{Node-wise contraction}

For one constrained node, let
$$
W_i=\sum_{e\rightarrow i}w_e,
\qquad
\bar z_i=\frac{\sum_{e\rightarrow i}w_ez_e}{W_i}.
$$
Weighted squared error decomposes as
$$
\sum_{e\rightarrow i}w_e\lVert Z_i-z_e\rVert^2
=
W_i\lVert Z_i-\bar z_i\rVert^2
+
\sum_{e\rightarrow i}w_e\lVert z_e-\bar z_i\rVert^2.
$$
The second term is fixed proposal disagreement.

For
$$
Z_{\mathrm c,i}=Z_i+\alpha R_i,
\qquad
\eta_i=\frac{\alpha W_i}{\epsilon+W_i},
$$
we obtain
$$
Z_{\mathrm c,i}-\bar z_i
=
(1-\eta_i)(Z_i-\bar z_i).
$$
The state-dependent term contracts by $(1-\eta_i)^2$
whenever $0\leq\alpha\leq1$.
Unconstrained nodes remain unchanged.

For negligible $\epsilon$ and $\alpha=0.25$, the contraction
factor is approximately $0.5625$.
Such a reduction follows from interpolation and provides
an implementation consistency check.

\subsection{Energy used by the implementation}

The code reports weighted mean per-sample energy:
$$
E_b(Z)=
\frac{
\sum_{e:b_e=b}w_e
\lVert z_e-Z_{b_e,i_e}\rVert^2/d
}{
\epsilon+\sum_{e:b_e=b}w_e
}.
$$
For fixed proposals and weights, its denominator is constant
with respect to $Z$.
Node-wise contraction therefore also implies non-increase
of this energy in exact arithmetic.

The acceptance test is
$$
E_b(Z_{\mathrm{candidate}})
\leq E_b(Z)+\delta.
$$
Its numerical guarantee includes the configured tolerance.
The full learned update can increase external energy.

\subsection{Prefix independence}

Causal convolutions and the attachment condition $i_e\geq a_e$
control local information flow.
They do not certify that external source states or support
decisions were constructed without target information.

The analytic consistency step with fixed scale is
position-local.
The supplied safeguard accepts scales using energy aggregated
across each sample.
If acceptance changes in response to suffix constraints,
the chosen scale can also alter earlier constrained nodes.

In exact arithmetic, the first admissible scale already
satisfies the contraction property.
A strict implementation-level prefix guarantee can use
the fixed analytic scale or separately establish
prefix-independent acceptance decisions.
The supplied causal test checks a particular suffix
perturbation rather than all acceptance patterns.

\subsection{Floating-point assembly}

Residuals and weights are accumulated in FP32 using
\texttt{index\_add\_}.
Permutation invariance is exact for the mathematical sum.
Numerical comparisons use tolerances because parallel
accumulation can change rounding order.

\subsection{Sufficient statistics of fixed observations}

For fixed adapted observations, define
$$
A_i=\sum_{e\rightarrow i}w_ez_e,
\qquad
W_i=\sum_{e\rightarrow i}w_e.
$$
Then
$$
R_i(Z)=\frac{A_i-W_iZ_i}{\epsilon+W_i}.
$$
This identity exposes the assembly's information boundary:
the consistency residual uses the weighted first moment
and total weight at each node.

Different observation sets can have the same $A_i$ and $W_i$.
Their residuals are then identical, even if their internal
disagreement differs.
That disagreement appears in the energy floor through
weighted second moments.
Consequently, summation is not a lossless representation
of the individual proposals.

These identities describe the implemented operator.
Precomputing moments or exposing disagreement as an additional
feature would be separate implementation choices.
They are not assumed in the reported experiments.

\section{Memory Representation, Publication, and Audit}
\label{app:memory}

\subsection{Storage units}

A source document maps to a published \emph{element version}.
An element can contain many 64-token \emph{blocks}.
Each block has a \emph{cell}, and decoding exposes
position-specific \emph{slots}.
Element counts, cell counts, and floating-value counts
are different quantities.

The production run contains 52,809 active element versions
near 1.7 billion stored floating-point values.
Stored parent states contribute to the physical budget.

\subsection{Codec and worker settings}

The codec production specification uses hidden and memory
dimensions 768, six decoder operators, kernel seven,
and an FFN multiplier of 2.5 rounded to a multiple of 256.
Tile span is 64 and tile width 48.
Optional parent context uses a coarse ratio of four.

Shared codec training alternates local-state adaptation
and shared-parameter updates.
The reported specification uses 32 local steps during
training episodes and 128 during validation.

Production workers freeze the codec.
Reported production invocations use 1,024 adaptation steps,
minimum token recall 0.70, and maximum reconstruction loss
2.60.
These are run settings, distinct from the defaults in
the supplied worker CLI.

\subsection{Published and local state}

The worker stores final memory cells in BF16 and adds
configured coarse states.
Metadata records codec identity and checksum, document
identity, source-token checksum, adaptation settings,
and reconstruction metrics.

The local record additionally stores initial cells,
residuals, final cells, and the adaptation curve.
It may be deleted after publication.
Source checksums identify the token sequence; they do not
measure semantic preservation.

\subsection{Registry semantics}

The journal supports READY, ACTIVATE, DEACTIVATE, RETIRE,
and SNAPSHOT events.
Materialization replays them into published, active,
retired, and snapshot records.
Activation requires an existing, non-retired version.
Replacing an active version changes registry state without
overwriting the previous artifact.

Retirement prevents selection in new active-set snapshots.
An existing snapshot remains a historical manifest, and
loading it does not inherently implement retroactive
revocation.
Deletion and access policies require separate treatment.

The publication protocol orders file completion before READY.
A crash after rename but before journal append can leave
an orphan file, which verification reports.
The registry is a shared-filesystem, lock-based implementation,
not a replicated consensus service.

Initialization should precede concurrent worker operation.
A truncated journal tail can be tolerated during reading,
but continued publication after a damaged tail requires
recovery handling.
Filesystem locking and durability must be checked in the
target local or NFS environment.

\subsection{Readout-cache audit}

\begin{table}[H]
\centering
\small
\caption{Audit of a selected frozen-codec readout cache.}
\label{tab:cache}
\begin{tabular}{lr}
\toprule
Quantity & Value \\
\midrule
Unique evidence blocks & 2,910 \\
Elements represented & 98 \\
QA task references & 10,000 \\
Valid source tokens & 186,031 \\
Exactly reconstructed tokens & 139,468 \\
Token reconstruction accuracy & 74.97\% \\
Completely reconstructed blocks & 1 \\
Completely reconstructed block fraction & 0.0344\% \\
Cache shards & 12 \\
Tensors and records & 300,395,921 bytes \\
Measured cache-build stage & 13.78 seconds \\
\bottomrule
\end{tabular}
\end{table}

The measured build stage excludes some startup work.
The cache passed the reported count and checksum checks.
Reconstruction quality is separate from file integrity.

\begin{table}[H]
\centering
\small
\caption{Representative cache schema for $N$ blocks.}
\label{tab:cache_schema}
\begin{tabular}{lll}
\toprule
Tensor & Shape & Dtype \\
\midrule
\texttt{block\_key\_hash} & $N$ & int64 \\
\texttt{memory\_cell} & $N\times768$ & float32 \\
\texttt{coarse\_parent\_cell} & $N\times768$ & float32 \\
\texttt{codec\_slot\_hidden} & $N\times64\times768$ & BF16 \\
\texttt{source\_token\_ids} & $N\times64$ & int64 \\
\texttt{source\_valid\_mask} & $N\times64$ & int32 \\
\texttt{decoded\_token\_ids} & $N\times64$ & int64 \\
\texttt{decoded\_top1\_logprob} & $N\times64$ & float32 \\
\texttt{decoded\_margin} & $N\times64$ & float32 \\
\bottomrule
\end{tabular}
\end{table}

Source IDs support auditing and source-based controls.
Their presence does not imply that every readout consumes
them.
Fields used for cue construction, support generation,
and neural input must be distinguished.

\section{Publication Lifecycle and Failure Boundaries}
\label{app:lifecycle}

This section records the operational consequences of the
registry protocol.
It is a protocol analysis, not a report of a completed
fault-injection benchmark.

\subsection{Three independent decisions}

A document can be successfully optimized without being
published.
A published version can remain inactive.
An active version can be absent from a consumer's existing
snapshot.
These are distinct states:

\begin{enumerate}
    \item \textbf{Optimization acceptance:} reconstruction
    satisfies the worker's publication threshold.
    \item \textbf{Registry visibility:} READY records the
    completed artifact.
    \item \textbf{Consumer visibility:} activation and
    snapshot selection determine use.
\end{enumerate}

This separation prevents a worker's local progress from
silently mutating a running consumer's evidence set.

\subsection{Failure windows}

\begin{table}[H]
\centering
\small
\caption{
Publication failure windows and their interpretation.
Recovery actions are operational requirements, not automatic
repair implemented by every code path.
}
\begin{tabular}{p{0.24\linewidth}p{0.34\linewidth}p{0.32\linewidth}}
\toprule
Interruption point & Observable state & Interpretation \\
\midrule
Before final artifact rename &
Temporary worker or upload file; no READY &
Incomplete work is not a published version. \\

After rename, before READY &
Final artifact path without journal publication &
Orphan file requiring inspection before version reuse. \\

After READY, before ACTIVATE &
Published version without active selection &
Artifact is available for explicit activation. \\

After ACTIVATE, before new snapshot &
Live active set has changed; old snapshots unchanged &
Existing consumers retain their previous selection. \\

After snapshot write, before SNAPSHOT event &
Self-contained snapshot file may lack its journal record &
File integrity and journal bookkeeping are separate checks. \\
\bottomrule
\end{tabular}
\end{table}

The exclusive lock prevents cooperating publishers from
claiming the same version concurrently.
It does not turn multiple filesystem and journal operations
into one indivisible transaction.
The ordering is designed so that a completed READY publication
refers to an already written file, subject to filesystem
durability assumptions.

\subsection{Replacement and rollback}

An update creates a new immutable version.
Activation changes the selected version for an element.
A new snapshot records that selection.
Returning to an older compatible snapshot restores its
external artifact selection without undoing neural optimizer
updates.

Behavioral rollback additionally depends on unchanged model
weights, tokenizer, decoding configuration, and deterministic
execution conditions.
The registry's ability to select an older artifact set is
distinct from an empirical guarantee of bitwise output equality
across hardware and software environments.

\subsection{Dependency identity}

A reproducible consumer configuration binds at least:
\begin{itemize}
    \item core and adapter checkpoint;
    \item tokenizer and input-code convention;
    \item codec configuration and weights;
    \item memory snapshot;
    \item retrieval index and candidate policy;
    \item readout-cache construction;
    \item VM implementation, configuration, and compiler;
    \item output scoring and stopping rules.
\end{itemize}

Changing a document version changes one dependency.
Changing a codec can invalidate derived readout caches
and adapter compatibility.
Changing a tokenizer can alter both lexical search keys
and render-buffer identities.
The dependency record therefore matters beyond file hashes
alone.

\section{Retrieval and Positional Readout}
\label{app:readout}

\subsection{Separate search charts}

The fixed index uses complementary views:
a positional prefix, sliding windows, an order-independent
token bag, and local $n$-grams.
Candidate outputs are merged before reranking.
The reported pipeline uses top-64 candidates followed by
a smaller attachment set.

A prefix sketch preserves one coordinate system.
An internal quotation can violate that alignment even when
its tokens occur verbatim.
Window and position-independent charts supply alternative
supports without changing memory cells.

Training supports candidate-miss policies
\texttt{error}, \texttt{skip}, and \texttt{inject}.
Candidate recall is measured before oracle insertion.
Injected-positive training is distinct from unmodified
retrieval evaluation.

\subsection{Memory readout controls}

The diagnostic interfaces are:
\begin{itemize}
    \item \texttt{dense}: one cell repeated over answer positions;
    \item \texttt{source\_positional}: fixed observations
    of original source-token identities;
    \item \texttt{codec\_positional}: hidden slot states
    of the frozen decoder;
    \item \texttt{gold}: oracle target-token observations.
\end{itemize}

\begin{table}[H]
\centering
\small
\caption{
Developmental memory-readout diagnostics.
Accuracy and greedy exact match are percentages.
Intervention columns are increases in NLL relative to the
corresponding unmodified condition.
These are diagnostic point estimates, not a pooled
multi-seed result.
}
\label{tab:readout}
\begin{tabular}{lrrrrr}
\toprule
Interface & NLL & Token acc. & Greedy EM &
$\Delta L_{\mathrm{wrong}}$ &
$\Delta L_{\mathrm{perm}}$ \\
\midrule
Dense cell & 6.6331 & 12.15 & 0.00 & 0.02 & 0.00 \\
Source positional & 5.5522 & 30.32 & 1.56 & 3.12 & 2.85 \\
Codec positional & 4.4729 & 48.60 & 7.81 & 5.52 & 4.87 \\
Gold positional & 4.2934 & 55.41 & 9.38 & 5.91 & 5.34 \\
\bottomrule
\end{tabular}
\end{table}

The measurements distinguish positional accessibility
from successful free serialization.
The codec and source-code interfaces provide different
feature spaces.
Their comparison does not isolate an intrinsic advantage
of latent storage over original text.

\subsection{Coordinate provenance}

A positional observation can contain token identity,
source slot $s$, answer-relative position $i$, length $L$,
and boundary flags:
$$
u_i=\operatorname{Encode}
\left(c_{\mathrm{tok}}(x_s),c_{\mathrm{src}}(s),
c_{\mathrm{ans}}(i),c_{\mathrm{len}}(L),
f_{\mathrm{first}},f_{\mathrm{last}}\right).
$$

In coordinate-assisted diagnostics, selection, alignment,
or length can come from the task record.
Question-only operation must derive them from the request
and available evidence.
These are separate conditioning regimes.

Integer fields require boundary conventions:
a six-bit field directly represents zero through 63,
so a length of 64 requires an offset or another representation.

\subsection{Specificity and availability}

Two paired differences are
$$
\Delta L_{\mathrm{availability}}
=L_{\mathrm{absent}}-L_{\mathrm{correct}},
$$
$$
\Delta L_{\mathrm{specificity}}
=L_{\mathrm{wrong}}-L_{\mathrm{correct}}.
$$
Availability includes any effect of enabling the external
path.
Specificity compares correct and wrong content.
Negative sampling and feature-statistic matching affect
the interpretation.

\subsection{Oracle-memory R1}

The supplied base trainer prolongs a source cell over the
following token block.
Wrong-document cells are resized to the required count,
and shuffled controls permute cells before prolongation.

Attachment timing does not prove that an offline cell
depends only on the preceding source prefix.
Document-wide adaptation and parent context can introduce
broader source dependence.
R1 is therefore an external-memory conditioning task.

Short source sequences may be repeated to fill the configured
length; metadata records this wrapping.
R1 validation samples from the configured snapshot.
A separate held-out element split is not imposed by that
base trainer itself.

\section{Autoregressive Readout and Coordinate Contracts}
\label{app:generation}

\subsection{Prediction positions}

The training convention uses externally shifted targets:
$$
\text{input}_i=x_i,
\qquad
\text{label}_i=x_{i+1}.
$$
The hidden state at position $i$ therefore predicts the
token at position $i+1$.
An observation intended to influence the first answer token
must be available at its predicting node, not merely at
the node occupied by that answer token during teacher forcing.

This distinction applies equally to source-slot observations,
VM returns, and EOS supervision.
A consistent interface records the relation between:
\begin{itemize}
    \item source position;
    \item input position;
    \item predicted-token position;
    \item answer-relative position;
    \item observation availability.
\end{itemize}

\subsection{Historical attachments}

The reference generation function asks an element provider
for observations at each prefix.
The backbone then recomputes the current field from token
coordinates and those observations.

If an observation affected an earlier node during training,
omitting it during prefix recomputation changes the input
to the recurrent solver.
Thus supplying only the newest attachment is not generally
equivalent to supplying the full still-relevant attachment
history.

The protocol can deliberately use either behavior, but it
must match the intended training and evaluation semantics.
The paper's rendering results retain the distinctions
documented for their respective implementations.

\subsection{Sliding-window coordinates}

When the prefix exceeds the model context, generation uses
a cropped window.
If $o$ tokens have been removed, an absolute target index
$i_{\mathrm{abs}}$ corresponds to
$$
i_{\mathrm{local}}=i_{\mathrm{abs}}-o.
$$
Availability coordinates require the same frame convention.
Observations outside the current window must be discarded,
re-expressed, or otherwise handled by a specified policy.

A provider that receives only the cropped token prefix
cannot always recover absolute document or answer offsets
from token content alone.
Absolute-position bookkeeping must therefore be retained
externally when the task requires it.

\subsection{Termination}

Three output protocols must be distinguished:
\begin{enumerate}
    \item an externally declared token count;
    \item model-predicted EOS or another stop token;
    \item a fixed maximum generation length.
\end{enumerate}

The VM rendering experiments use the first.
Strict free-generation evaluation should additionally
measure termination correctness and full output equality.
Extracting the first signed integer answers a narrower
question than exact serialization of the entire response.

\subsection{Readout validation sequence}

A practical validation sequence is:
\begin{enumerate}
    \item verify a single externally specified token at its
    predicting node;
    \item verify a short random token sequence with known
    supports and length;
    \item compare teacher-forced and prefix-recomputed logits;
    \item test retained historical attachments;
    \item test window-boundary remapping;
    \item replace oracle supports with inferred supports.
\end{enumerate}

This is a recommended diagnostic progression, not an additional
set of measured results.
It separates transmission errors from source selection
and content reconstruction.

\section{Executable Skills and Replay Semantics}
\label{app:vm}

\subsection{Machine configuration}

The reference configuration provides 16 registers,
256 memory cells, signed 64-bit integers, maximum program
length 256, execution limit 1,024 steps, and output limit
4,096 characters.

Instructions cover loads, transfers, bounded memory,
arithmetic, bitwise operations, shifts, comparisons,
absolute jumps, integer and character output, and halting.
The opcode count is obtained from the implementation's
enumeration.

Signed wrapping is
$$
u=x\bmod2^w,
\qquad
\operatorname{wrap}_w(x)=
\begin{cases}
u,&u<2^{w-1},\\
u-2^w,&u\geq2^{w-1}.
\end{cases}
$$
Division truncates toward zero, with modulo
$$
q=\operatorname{trunc}(a/b),
\qquad r=a-qb.
$$
Right shift is arithmetic.
Jump targets are validated even when a conditional jump
is not taken.
Character output excludes surrogate code points.

\subsection{Fault and parsing boundaries}

Runtime statuses distinguish normal halt, execution and
program limits, invalid instructions or arguments,
out-of-bounds access, division by zero, output limits,
and invalid program counters.

Schema parsing and configuration validation can raise
host-language exceptions.
The typed-runtime-fault contract applies to inputs reaching
the executor through the supported schema.
Instruction bounds do not alone bound arbitrary input-file
or metadata size.

The trace records pre- and post-instruction registers,
comparison flags, PCs, memory writes, and output fields.
An attempted output can appear in \texttt{output\_append}
even when an output-limit fault prevents commitment.
Consumers should check status and final output rather than
interpret that field alone as a committed-write log.

\subsection{Replay identity}

The program checksum covers canonical serialization,
including metadata.
The result checksum covers terminal status, state, output,
trace, and program checksum.

The supplied result hash does not include the VM configuration
as a separate field.
A complete replay manifest must retain configuration and
implementation identity.
The \texttt{verify-replay} CLI uses the default configuration;
nondefault replay requires constructing the corresponding
configuration explicitly.

\subsection{Arithmetic and loop programs}

The arithmetic builder emits five instructions:
load two immediates, apply the operation, output the result,
and halt.
The factorial builder supplies a standalone loop example.
This is separate from learned arithmetic routing.

The included self-tests cover basic arithmetic, negative
division and modulo, factorial control flow, division-by-zero
status, step limits, memory access, replay, and signed wrapping.
Test definitions and executed test reports are separate artifacts.

\subsection{Routing and positional return}

An early prompt placed routing after variable-length operands.
For a local causal model, longer tokenizations can move
the operation cue outside the routing state's receptive field.
The revised prompt places routing next to that cue.

For result $v$ with tokens $r_0,\ldots,r_{M-1}$,
$$
u_i^{\mathrm{VM}}=
\left[
c_{\mathrm{val}}(v)\Vert c_{\mathrm{tok}}(r_i)
\Vert c_{\mathrm{pos}}(i)\Vert c_{\mathrm{len}}(M)
\right].
$$
Character channels are optional.
The reported renderer uses $M$ as generation length and
extracts the first signed integer for numeric exact match.

Substitutions are evaluated separately on changed cases
$\widetilde v\neq v$.
Unchanged returns cannot distinguish following the supplied
value from preserving the original answer.

\subsection{Oracle-program S1.1}

The base trainer supervises an instruction scaffold and
receives the oracle VM result.
Its bounded argument labels use modular classes, which
cannot uniquely encode integers outside the class range.
Later restricted routing separates deterministic operand
extraction from learned operation selection.

In the supplied S1.1 builder, the return is attached after
the last controller slot, whereas the first answer token
is predicted at the preceding position under external
label shifting.
This alignment belongs to the early protocol, not the
later positional renderer.

\section{Configurations, Training, and Instrumentation}
\label{app:training}

\subsection{Architecture ledger}

\begin{table}[H]
\centering
\small
\caption{Reported architecture specifications.}
\begin{tabular}{p{0.27\linewidth}p{0.65\linewidth}}
\toprule
Component & Configuration \\
\midrule
Multi-Mesh large run &
Width 2,048; five parameterized stages; two iterations
per stage; three coarse levels; ratio four; tile span 64;
fine/coarse/scratch kernels 7/5/3; FFN multiplier 4.25;
tile width 64; tile-state dimension 2,048. \\

Modular pilot &
Width 512; context 512; four shared iterations;
local/correction kernels 7/3; FFN multiplier two;
memory input 768; VM input 128; update scale 0.25. \\

Codec production &
Hidden and memory dimensions 768; six decoder operators;
kernel seven; FFN multiplier 2.5 rounded to a multiple
of 256; tile span 64; tile width 48. \\
\bottomrule
\end{tabular}
\end{table}

The selected-evidence release uses a larger modular core
specified by its artifact configuration.
Pilot settings are not substituted for that checkpoint.

\subsection{Training conventions}

Large-run specifications use FP32 parameters, BF16 autocast,
AdamW, externally shifted labels, and gradient clipping.
The reference Transformer launch uses two GPUs,
microbatch eight per GPU, eight accumulation steps,
and context 2,048:
$$
2\cdot8\cdot8\cdot2048=262{,}144
$$
prediction targets per update.

The specified peak learning rate is $1.5\cdot10^{-4}$,
minimum $10^{-5}$, warmup 2,000 updates, weight decay
0.01, Adam coefficients 0.9 and 0.95, and clipping 1.0.
The Multi-Mesh command targets the same token batch
with microbatch one and 64 accumulation steps per GPU.
Its initial correction scale is 0.18.
Invocation and checkpoint records determine actual settings
and realized budgets.

\subsection{Joint objectives}

Integration uses language-only, memory-conditioned,
and skill-conditioned examples:
$$
L=\lambda_0L_{\mathrm{R0}}+
\lambda_2L_{\mathrm{R2}}+
\lambda_{\mathrm{ans}}L_{\mathrm{S2,answer}}+
\lambda_{\mathrm{route}}L_{\mathrm{route}}.
$$
Joint updates provide contemporaneous supervision to
the core and interface heads.
The earlier R0/R1/S1.1 trainer instead samples curriculum
modes over microsteps.
These are distinct implementations.

\subsection{Task weights and supervised-token weights}

Let $\mathcal I_g$ be the supervised positions of task
group $g$, and let $\lambda_g$ be its desired coefficient.
Uniform within-group weights can be defined as
$$
\omega_{b,i}=\frac{\lambda_g}{|\mathcal I_g|},
\qquad (b,i)\in\mathcal I_g.
$$
Then
$$
L_{\mathrm{token}}
=
\sum_g\sum_{(b,i)\in\mathcal I_g}
\omega_{b,i}\ell_{b,i}
=
\sum_g\lambda_g
\frac{1}{|\mathcal I_g|}
\sum_{(b,i)\in\mathcal I_g}\ell_{b,i}.
$$

The model's weighted-loss path sums weighted token losses
without dividing again by the total weight.
This permits each group's weights to encode its requested
coefficient.
Padding and ignored labels contribute zero effective weight.

\begin{algorithm}[H]
\caption{Task-normalized joint objective}
\label{alg:joint}
\begin{algorithmic}[1]
\Require Task groups, labels, masks, task coefficients;
controller targets where applicable
\Ensure Scalar objective for one optimizer update
\State Build each group's inputs and external observations
\State Mark supervised target positions and ignored positions
\For{each nonempty group $g$}
    \State Assign supervised weights summing to $\lambda_g$
\EndFor
\State Evaluate token losses with external label alignment
\State $L_{\mathrm{token}}\gets
\sum_{b,i}\omega_{b,i}\ell_{b,i}$
\State Compute controller loss on declared routing positions
\State $L\gets L_{\mathrm{token}}+
\lambda_{\mathrm{route}}L_{\mathrm{route}}$
\State Backpropagate through core, adapters, and controller
\State Keep published memory states and VM execution fixed
\State \Return $L$
\end{algorithmic}
\end{algorithm}

This algorithm states the weighting convention supported
by the model's joint-loss interface.
The concrete grouping, coefficients, and batching follow
the integration trainer's configuration.

Changing group sequence length without preserving its
weight normalization changes the effective objective.
Likewise, a large count of padding positions is not a
large count of supervised prediction targets.
Both physical positions and supervised targets should
be recorded.

\subsection{Multi-Mesh diagnostics}

The trainer records:
\begin{itemize}
    \item RMS, extrema, finite fractions, and storage size
    of fine, coarse, scratch, and tile states;
    \item relative updates and old/new cosine similarity;
    \item local, coarse, scratch, and tile transfer magnitudes;
    \item module-group gradient norms;
    \item logit changes under transfer-path interventions;
    \item forward timing and allocated/reserved memory.
\end{itemize}

Relative update is
$$
\rho=
\frac{\operatorname{RMS}(H^{k+1}-H^k)}
{\operatorname{RMS}(H^k)+\epsilon}.
$$
A transfer-to-total RMS ratio is relative amplitude,
not an additive percentage of causal contribution.
Correlated or opposing components need not sum in RMS.

Startup checks cover fixed input parameters, tile roundtrip,
mesh shapes, suffix perturbations, gradient flow, and
unused parameters.
Test length is chosen to activate completed cells at every
configured hierarchy level.

\subsection{Implementation boundaries}

The supplied Multi-Mesh checkpoint writer saves rank-zero
sampler and RNG state rather than a complete per-rank
continuation record.
Its wall-clock diagnostic triggers are computed locally
before collective operations.
Reproducible distributed continuation and synchronized
diagnostic scheduling require attention to these boundaries.

The modular code exposes plain-tensor training outputs
and explicit gradient checks.
Computing loss outside a forward method is valid in general;
failures depend on the actual wrapper and distributed graph.

Large finite activations alone do not identify the first
nonfinite operation.
Diagnosis should distinguish activation growth, reductions,
products, and backward computation.

\section{Verification Matrix and Experimental Boundaries}
\label{app:verification}

Verification is organized by the property being checked,
not by a single pass/fail label for the entire system.
Table~\ref{tab:verification} summarizes checks represented
in the supplied code and their scope.

\begin{table}[H]
\centering
\small
\caption{
Implementation checks and what they establish.
The table describes test coverage, not an assertion that
all configurations were tested on every platform.
}
\label{tab:verification}
\begin{tabular}{p{0.25\linewidth}p{0.34\linewidth}p{0.31\linewidth}}
\toprule
Check & Tested property & Boundary \\
\midrule
Fixed input &
No trainable input parameters; codebook does not require gradients &
Does not reconstruct historical optimizer activity. \\

Assembly permutation &
Reordered observations preserve logits within tolerance &
Finite precision and tested inputs only. \\

Consistency trace &
Accepted consistency substep respects energy tolerance &
Excludes the full neural correction. \\

Suffix perturbation &
Earlier logits remain unchanged in a specified case &
Not exhaustive over support or acceptance patterns. \\

Variable element count &
Finite outputs with and without observations; outputs can differ &
Does not establish correct use of arbitrary elements. \\

Gradient flow &
Each tested module group receives finite nonzero gradient &
Not a claim that every parameter is useful. \\

Serialization &
Restored weights preserve tested outputs &
Depends on implementation and numerical environment. \\

VM replay &
Repeated execution reproduces the result checksum &
Configuration must be retained separately. \\
\bottomrule
\end{tabular}
\end{table}

\subsection{Layered correctness}

The architecture has several independent notions of correctness:

\begin{enumerate}
    \item \textbf{Artifact integrity:} stored bytes match their
    recorded checksum.
    \item \textbf{Representation fidelity:} a memory preserves
    the relevant source content.
    \item \textbf{Execution correctness:} the VM follows the
    intended program under specified semantics.
    \item \textbf{Selection correctness:} the selected evidence
    or operation is appropriate for the request.
    \item \textbf{Interface correctness:} observations are aligned
    with their intended neural targets.
    \item \textbf{Answer correctness:} the final output satisfies
    the task.
\end{enumerate}

A successful check at one layer does not establish all later
layers.
Conversely, an answer error does not identify which earlier
layer failed.
The modular design exposes these boundaries for measurement.

\subsection{Failure localization}

A useful diagnostic order is:
\begin{enumerate}
    \item replay the external component independently;
    \item verify the selected artifact and source identifiers;
    \item inspect the emitted observations and supports;
    \item compare correct, absent, and substituted observations;
    \item compare teacher-forced and generated-token paths;
    \item inspect final formatting and stopping.
\end{enumerate}

This procedure distinguishes unavailable information from
information that is present but not successfully transmitted.
It also distinguishes execution failure from output rendering
failure, a separation used directly in the arithmetic studies.

\subsection{Evidence needed for stronger claims}

Different claims require different tests.
Independent production is established by the worker and
publication architecture.
Behavioral benefit requires task evaluation.
Stable attachment across growing stores requires a fixed old
query set as well as newly answerable queries.
Serving scalability requires full-store measurements rather
than selected-evidence packages.
These extensions can use the same interface while evaluating
different system properties.

\section{Resource Accounting and Serving}
\label{app:resources}

Resources are divided into trainable parameters, frozen
shared weights, persistent external state, derived indices,
readout caches, and per-query active state.

A 768-value BF16 cell occupies 1,536 bytes.
A 64-token uint16 source block occupies 128 bytes.
A BF16 decoder-slot cache occupies
$$
64\cdot768\cdot2=98{,}304
$$
bytes per block before other records.

\begin{table}[H]
\centering
\small
\caption{Resource ledger.}
\begin{tabular}{p{0.28\linewidth}p{0.64\linewidth}}
\toprule
Category & Contents \\
\midrule
Trainable parameters &
Core, adapters, controller, and reranker. \\
Frozen shared weights &
Codec and other shared frozen transforms. \\
External state &
Final cells, coarse states, metadata, and version records. \\
Search state &
Chart keys, lookup structures, and candidate metadata. \\
Readout cache &
Slot states, decoded tokens, and diagnostic fields. \\
Active query state &
Selected observations, neural activations, and VM trace. \\
Execution cost &
Retrieval, loading, decoding, assembly, neural processing,
and generation. \\
\bottomrule
\end{tabular}
\end{table}

\subsection{Serving paths}

On-demand decoding lowers persistent cache requirements
and adds request-time decoder work.
Cached readout moves work into preprocessing while
increasing storage and transfer.

At fixed type set, adapter parameters do not grow with
document count.
Storage and indexing do grow, and per-query cost depends
on selected observations.
Independent production does not imply constant total cost.

\subsection{Stage-wise latency}

An end-to-end measurement should distinguish
$$
t_{\mathrm{query}}=
t_{\mathrm{retrieve}}+
t_{\mathrm{load}}+
t_{\mathrm{decode}}+
t_{\mathrm{assemble}}+
t_{\mathrm{core}}+
t_{\mathrm{render}},
$$
with overlap reported where asynchronous execution makes
the simple sum inapplicable.

Cache hits, cold loads, and filesystem effects change the
measurement.
Decoder timing should identify whether parent states and
slot caches are already resident.
GPU timing requires synchronization appropriate to the
measured interval.

A precomputed selected-evidence package omits parts of
this path.
Its latency cannot be substituted for full-store retrieval
and loading latency.

\subsection{Construction cost and amortization}

Memory construction incurs local optimization cost before
serving.
A full comparison should therefore report both offline
construction and online use.

For a collection used by $Q$ requests, an accounting quantity
is
$$
C_{\mathrm{total}}(Q)=
C_{\mathrm{construct}}+
C_{\mathrm{index}}+
C_{\mathrm{cache}}+
\sum_{q=1}^{Q}C_{\mathrm{serve},q}.
$$
This is a bookkeeping definition, not a measured scaling law.
The relevant trade-off depends on update frequency,
reuse, quality requirements, and the alternative system.

Raw-text storage has a different construction cost and
may require a different reader.
A fair comparison should include the complete reader path
rather than comparing only cell bytes or core parameter counts.

\section{Fixed Observations and Visual Resources}
\label{app:visual}

Binary16 supplies fixed token identities for vocabularies
of at most $2^{16}$ entries.
The input lift is parameter-free.
A materialized codebook is a frozen buffer.

The Multi-Mesh tile packs
$$
64\cdot16=1024=32\cdot32
$$
identity bits.
Visual-form and boundary channels supplement this identity
representation.

The visual resource is GNU Unifont 14.0.01
\citep{unifont14001}.
The reported pipeline uses NFC normalization, specified
whitespace/control handling, font size 64, 64-pixel
character cells, area downsampling, threshold 192,
and at most 32 normalized characters per token piece.

Visual similarity does not define token identity.
The release retains upstream font notices and separately
documents code, model, and resource licensing.
GNU Unifont releases from version 13.0.04 provide licensing
under SIL OFL 1.1 and GPL version 2 or later with the
GNU font embedding exception.

\section{Evaluation Provenance}
\label{app:evaluation}

Each result is interpreted through its checkpoint,
conditioning regime, data split, and scoring rule.
The relevant manifest fields include:
\begin{itemize}
    \item checkpoint, configuration, and code identity;
    \item dataset/task revisions and evaluated sample counts;
    \item prompt, BOS/EOS, context, truncation, and label alignment;
    \item codec and memory snapshot identity;
    \item cue, candidate, and support-construction policy;
    \item generation limits, termination, and normalization;
    \item token, numeric, or whole-answer metric definitions.
\end{itemize}

Tasks sharing an element or block are statistically dependent.
Resampling should preserve that grouping.
Paired controls use identical records and scoring rules.

Selected-evidence attachment, coordinate-assisted retrieval,
oracle readout, and unrestricted question-only operation
are separate protocols.
Their distinction identifies what each experiment measures.

\subsection{Checkpoint and run identity}

A launch command specifies intended settings.
A checkpoint records attained optimizer and token counts.
A model export identifies the tensors consumed by evaluation.
These records should be linked rather than treated as
interchangeable descriptions.

For interrupted training, the realized data stream also
depends on sampler and RNG restoration.
Matching a nominal seed and recipe does not establish
identical sampled windows across implementations or resumes.

\subsection{Comparison units}

Standalone benchmarks compare checkpoint behavior.
Memory reconstruction compares decoded positions with
source tokens.
Retrieval evaluates element or block selection.
Attachment compares evidence conditions under fixed weights.
Execution-result rendering compares neural adoption of
externally computed values.

Reporting these units separately prevents token accuracy
from being interpreted as answer exact match, candidate
recall as QA success, or stored values as trainable
model parameters.

\section{Artifacts}
\label{app:artifacts}

The released artifacts include the standalone prototype,
the selected-evidence attachment demonstration, and the
same-corpus Transformer reference:

\begin{itemize}
    \item \textbf{Standalone Multi-Mesh LM:}\\
    \url{https://huggingface.co/Bochkov/fem-multi-mesh-1p7b}

    \item \textbf{Selected-evidence modular attachment demonstration:}\\
    \url{https://huggingface.co/Bochkov/modular-reasoning-0p5b-g6p5-demo}

    \item \textbf{Same-corpus Transformer reference,
    100B-token training budget:}\\
    \url{https://huggingface.co/Bochkov/ab_ext_learned}
\end{itemize}

The supplementary source includes:
\begin{itemize}
    \item \texttt{reasoning\_model.py}: typed observations,
    adapters, assembly, recurrent correction, and unit tests;
    \item \texttt{reasoning\_trainer.py}: R0/R1/S1.1
    conditioning and curriculum;
    \item \texttt{memory\_worker\_v2.py}: local adaptation
    and publication;
    \item \texttt{memory\_registry.py}: append-only lifecycle
    and immutable snapshots;
    \item \texttt{skill\_vm.py}: bounded integer execution,
    trace serialization, replay, and self-tests;
    \item \texttt{trainer\_fem\_multi\_mesh.py}:
    standalone training and mesh diagnostics.
\end{itemize}

Additional research components implement codec training,
QA generation, multi-chart retrieval, positional readout,
and later execution-result protocols.
Their dataset and artifact dependencies are documented
separately from the runnable attachment package.

Repository locations identify distributions.
Specific results depend on evaluated revisions,
configurations, and weight files.

\end{document}